# A Deep Reinforcement Learning Based Multi-Criteria Decision Support System for Textile Manufacturing Process Optimization

ZHENGLEI HE  KIM-PHUC TRAN  SEBASTIEN THOMASSEY  XIANYI ZENG

*ENSAIT, GEMTEX – Laboratoire de Génie et Matériaux Textiles, F-59000 Lille, France*

JIE XU, CHANGHAI YI

*Wuhan Textile University, 1st, Av Yangguang, 430200, Wuhan, China*

*National Local Joint Engineering Laboratory for Advanced Textile Processing and Clean Production, 430200, Wuhan, China*

Textile manufacturing is a typical traditional industry involving high complexity in interconnected processes with limited capacity on the application of modern technologies. Decision-making in this domain generally takes multiple criteria into consideration, which usually arouses more complexity. To address this issue, the present paper proposes a decision support system that combines the intelligent data-based models of random forest (RF) and a human knowledge-based multi-criteria structure of analytical hierarchical process (AHP) in accordance with the objective and the subjective factors of the textile manufacturing process. More importantly, the textile manufacturing process is described as the Markov decision process (MDP) paradigm, and a deep reinforcement learning scheme, the Deep Q-networks (DQN), is employed to optimize it. The effectiveness of this system has been validated in a case study of optimizing a textile ozonation process, showing that it can better master the challenging decision-making tasks in textile manufacturing processes.

## 1. Introduction

There has been globally increasing competition in the textile industry that forces the manufacturers to innovatively promote the product quality, process efficiency, and process environmental issues as a whole. Compared with the other industrial sectors, textile manufacturing is a traditional sector relied on small and medium enterprises in general and it is highly complex due to the intricate relationship involved in a large number of parameter variables from a variety of processes[1]. It is nearly impossible to upgrade the textile manufacturing processes directly by only following the cases from other industries without considering the detailed characteristics of this sector and specific investigations in the applicable advanced technologies. To this end, the construction of accurate models for simulating manufacturing processes using intelligent techniques is rather necessary[2]. However, the decision space in a model relating to the different combinations of process and variable of textile manufacturing can be enormous and stochastic as there are numerous factors interactively impacting the performance[3]. Therefore, developing a decision support system to optimize the solutions in this issue remains an open challenge.

As the relationship between process parameters and product properties is not clearly known for textile manufacturing processes, the decision-maker is unaware of the probabilities of future states, which turns out that the present situation of decision making mostly is under uncertainty or risk[4]. By taking advantage of models learning from data and knowledge on the basis of artificial intelligence[5], a decision support system can make a difference in this regard. The applications of decision support systems for optimizing textile processes have been reported with various techniques: genetic algorithm [6] and fuzzy technology [7], [8] etc. But along with the development of artificial intelligence techniques and the growth of complexity in textile manufacturing, those classical approaches are no longer efficient in some scenarios. This is due to the fact that in recent years, a growing number of textile manufacturing problems were come up with large-scale data and high dimensional decision space[9], and instead of a single standard, multi-criteria is increasingly taken into consideration in these problems as evaluating the performance of a textile manufacturing[10].

Factors of textile manufacturing process consist of both objective and subjective effects, upon which this paper proposes a decision support system for optimizing the textile manufacturing process by combining the intelligent data-based models of random forest (RF) and human knowledge-based multi-criteria structure of analytical hierarchical process (AHP). Here, the proposal of the ensemble learning approach of RF for modeling textile process lies in the excellent approximation ability RF shown in a previous study[11] to deal with the complex and uncertain impacts of textile process variables on its performance, whereas the application of AHP, a multi-criteria decision making (MCDM) tool, regards to the fact that there are a few criteria govern the quality of textile process performance and their significance with an overall objective is different. Meanwhile, concerning the growing complexity in terms of large-scale data and high dimensional decision space in the textile manufacturing sector, this developed system formulates the textile manufacturing process optimization problem into a Markov decision process (MDP) paradigm and applies deep reinforcement learning (more specifically, the Deep Q-networks, DQN) instead of current methods to collaboratively approach the optimization problems in the textile manufacturing process.

The main contributions of this paper are listed below:

(1) The uncertainties of solutions on textile manufacturing performances are driven out by the RF approach in terms of process modeling.
(2) In terms of the widely existing MCDM problems in the field of textile, AHP is presented in this study to cooperatively work with the aforementioned measures.
(3) Formulation of the textile manufacturing process optimization as a Markov decision process problem and the solution based on the RL algorithm is proposed for the first time to deal with decision-making issues in the textile industry.
(4) The application of DQN is extended to support decision making on textile manufacturing solutions. Compared to the tabular RL algorithms applied in prior related works, DQN is more applicable and preferred to cope with the complicated realistic problem in the textile industry.
(5) Construction of a decision support system for textile manufacturing process optimization with case application.

The remainder of this paper is structured as follows: Section 2 summarizes relevant works. Section 3 describes in detail the study background in terms of problem formulation and the proposed algorithms for multi-criteria decision making in the presence of uncertainty. The detailed framework of the proposed system and a case study of applying the system to optimize an advanced textile finishing process optimization are introduced in Section 4 and Section 5 respectively. Finally, the discussions about future prospects and limitations are summarized in Section 6. Section 7 concludes this paper.

**2. Relevant literature**

Textile manufacturing originates from the fibers (e.g. cotton) to final products (such as curtain, garment, and composite) through a very long procedure with a wide range of different processes filled with a large number of variables. Decision-makers should be very knowledgeable about all the processes and parameters so that a solution of production scheme with optimal parameter setting could be provided quickly and properly from the numerous possibilities. This is extremely challenging with a dramatically high cost and is even impossible.

The decision support system can make a difference in this issue on the basis of intelligent techniques for modeling, optimization, and decision making. Majumdar et al. [12] briefly outlined the methods applied in the textile industry for process modeling (linear regression, artificial neural network, and fuzzy logic), optimization (linear programming, genetic algorithm, simulated annealing, Tabu search, and ant colony optimization) and multi-criteria decision making (AHP, and TOPSIS i.e. technique for order preference by similarity to ideal solutions ). Regarding the intelligent techniques, the combined use of artificial neural networks (ANN) and genetic algorithm (or evolution algorithm) was generally the first choice that researchers applied to deal with optimization problems in the textile domain [6], [13]–[17]. However, there are also numerous attempts were conducted on other methods, like adaptive network-based fuzzy inference system (ANFIS)[18], support vector machine [19]–[22], and gene expression programming [23], [24] etc. Due to the fact that the data in the textile industry is always limited for most of the processes nowadays, and our previous study released that the random forest (RF) model can work well in this situation compared with the extreme learning machine based ANN and SVM [11], so the RF model is proposed to simulate the textile manufacturing process in this paper, as a part of the proposed decision support system. In terms of the optimization techniques, it is known that the genetic algorithm and grey relational analysis [25] have been successfully applied in many other literatures for textile process decision-making, but in industry 4.0 era, the massive quantities of data and the high dimensional decision space of optimizing a textile manufacturing process may hinder the application of these tools. Reinforcement learning (RL) is a machine learning approach using a well understood and mathematically grounded framework of MDP that has been broadly applied to tackle the practical decision-making issues in the industry. For example, the pricing optimization[26]–[30], and the production or workflow scheduling [31], [32], as well as the energy management associated problems[33], [34]. Furthermore, using the temporal difference based RL methods to reduce the dimension of data in feature selection has been reported by Mehdi et al. [35] , and Jasmin et al[36] have applied the RL to approach the economic dispatch problem. Deep reinforcement learning (DRL) is an intelligent RL based technique that can well handle the large scale data and high-dimensional decision space. Related application of DRL for decision-making has been reported[37][38], however, at present, there is no complete study to solve a complex production problem, especially in the textile manufacturing industry.

As the most frequently used and widely discussed MCDM from the recently developed discipline of operation research, AHP has been proven to be an extremely useful decision-making method in the textile industry from the issued applications of AHP estimating the quality of fibers[39]and fabrics[40], the functional clothing design[12], rotor spinning machine setting[10], and the maintenance strategy evaluation[41], though certain reports have come up with their concerns on the theoretical basis of AHP[42]–[44]. The popular application and discussion of AHP in the textile industry are owing to its involvement of both objective and subjective factors that agree with the characteristic of the decision-making problem in the textile manufacturing process.

In summary, previous work addresses the optimization problems in the textile manufacturing process using methods different from the ones we are proposing in this article. Their approaches were found either performed not well relatively, or barely containing big data and high complexity. In the proposed framework, the RL would be cooperatively applied with RF models and AHP to optimize the solutions of the textile manufacturing process against multi-criteria.

## 3. Background

### 3.1. *Problem formulation*

Suggest a textile manufacturing process $P$ involves a set of parameter variables $\{v_1, v_2 \ldots v_n\}$, and the performance of this process is evaluated by multi-criteria of $\{c_1, c_2 \ldots c_m\}$. Decision making needs to figure out how those parameter variables affect the process performances in terms of each criterion, and whether a solution of $P$ $\{v_{1i}, v_{2j} \ldots v_{nk}\}$ is good or not relating to $\{c_1, c_2 \ldots c_m\}$.

Suppose there is a model maps $v_1, v_2 \ldots v_n$ of the process to its performance in accordance with $\{c_1, c_2 \ldots c_m\}$, then the performance of the specific solution could be presented by:

$$f_i(v_1, v_2 \ldots v_n) \mid c_i, \quad for \quad i = 1, \ldots m \tag{1}$$

When the domain of $v_j \in V_j$ is known, and the multi-criteria $\{c_1, c_2 \ldots c_m\}$ problem could be somehow represented by $C$, and the Equation (1) could be simplified to (2), and so that the objective of decision-makers is to find (3):

$$f(v_1, v_2 \ldots v_n) \mid C, \quad v_j \in V_j \tag{2}$$

$$argmax_{v_j \in V_j}[\, f(v_1, v_2 \ldots v_n) \mid C\,] \tag{3}$$

Equation (3) aims at searching the optimal solution of variable settings, while the traditional operation in this area usually relied heavily on trial and error.

### 3.2. *Random forest models*

The RF is a predictive model composed of a weighted combination of multiple regression trees to map the inputs and targets by learning from data. It constructs each tree using a different bootstrap sample of the data. It is different from the decision tree due to the splitting where each node uses the best split among all variables, by contrast, the RF uses the best among a subset of predictors randomly chosen at that node[45]. Generally, combining multiple regression trees increases predictive performance. It performs a prediction accurately by taking advantage of the interaction of variables and the evaluation of the significance of each variable[46]. The RF regressor is an ensemble-learning algorithm depending on a bagging method that combines multiple independently-constructed decision tree predictors to classify or predict certain output variables[46]. In the RF, successive trees do not rely on earlier trees; they are independent by using a bootstrap sample of the dataset, and therefore a simple unweighted average over the collection of grown trees $\{h(x,\Theta k)\}$ would be taken for prediction in the end. :

$$\bar{h}(x) = \frac{1}{K}\sum_{k=1}^{K} h(x, \Theta_k) \tag{4}$$

where $k=1,...,K$ is the number of trees, $x$ represents the observed input vector, $\Theta$ is an independent identically distributed random vector that the tree predictor takes on numerical values. RF algorithm starts from randomly drawing *ntree* bootstrap samples from the original data with replacement. And then grow a certain number of regression trees in accordance with the bootstrap samples. In each node of the regression tree, a number of the best split (*mtry*) randomly selected from all variables are considered for binary partitioning. The selection of the feature for node splitting from a random set of features decreases the correlation between different trees and thus the average prediction of multiple regression trees is expected to have lower variance than individual regression trees[47]. Regression tree hierarchically gives specific restriction or condition and it grows from the root node to the leaf node by splitting the data into partitions or branches according to the lowest Gini index:

$$I_G(t_{X(X_i)}) = 1 - \sum_{J=1}^{M} f(t_{X(X_i)}, j)^2 \tag{5}$$

where $f(t_{X(X_i)}, j)$ is the proportion of samples with the value $x_i$ belonging to leave $j$ as node $t$[48].

The application of random forest in the textile industry has been addressed to detect textile surface defects[49] and predict the photovoltaic properties of phenothiazine dyes[50]. According to our previous research[11], RF model was found that can well tackle the uncertainties in predict the unknown performance of different textile process solutions, therefore this study proposed RF models to construct the proposed framework. In the RF models, variables of the textile manufacturing process, $v_1, v_2... v_n$, constitute the input, while the targets, as described in Equation(1), are determined by the process performance against multi-criteria $\{c_1, c_2... c_m\}$.

### 3.3. *Analytic Hierarchy Process for Multi-criteria optimization*

The multi-criteria decision making (MCDM) problem presented in Equation (3) could be summarized as a single objective optimization problem by structuring a hierarchy of criteria in terms of weights or priorities:

$$argmax_{v_j \in V_j} [f(v_1, v_2 ... v_n) \mid C] = argmax_{v_j \in V_j} \sum_{i=1}^{m} w_i f_i(v_1, v_2 ... v_n) \tag{6}$$

where $w_1$ to $w_m$ are weights of criterion $c_1$ to $c_m$ respectively.

The AHP is a MCDM method introduced by Saaty [51] that uses a typical pair-wise comparison method to extract relative weights of criteria and alternative scores and turns a multi-criteria problem to the paradigm of Equation (6). Above all, it constructs a pairwise comparison matrix of attributes using a nine-point scale of relative importance, in which number 1 denotes an attribute compared to itself or with any other attribute as important as itself, the numbers of 2, 4, 6 and 8 indicate intermediate values between two adjacent judgments, whereas the numbers 3, 5, 7 and 9 correspond to comparative judgments of 'moderate importance', 'strong importance', 'very strong importance' and 'absolute importance' respectively. A typical comparison matrix ($C_m$) of $m \times m$ could be established for $m$ criteria as demonstrated below:

$$C_m = \begin{bmatrix} 1 & \cdots & a_{1m} \\ \vdots & \ddots & \vdots \\ a_{m1} & \cdots & 1 \end{bmatrix} \tag{7}$$

where $a_{ij}$ represents the relative importance of criterion $c_i$ regarding criterion $c_j$. Thus $a_{ij} = \frac{1}{a_{ji}}$ and $a_{ij} = 1$ when $i = j$. Note that a consistency index (*CI*) is introduced in AHP with consistency ratio (*CR*) on the basis of the principal eigenvector ($\lambda_{max}$) to validate the consistency in the pairwise comparison matrix:

$$CI = \frac{\lambda_{max} - m}{m - 1} \quad and \quad CR = \frac{CI}{RCI} \tag{8}$$

where *RCI* is a random consistency index and the values of it are available in[42]. Afterward, the relative weight of the $i_{th}$ criteria ($w_i$) would be calculated by the geometric mean of the principal eigenvector, $i_{th}$ row ($GM_i$), of the above matrix, and then normalizing the geometric means of rows:

$$GM_i = \left\{\prod_{j=1}^{m} a_{ij}\right\}^{\frac{1}{m}} \quad and \quad w_i = \frac{GM_i}{\sum_{I=1}^{m} GM_i} \tag{9}$$

### 3.4. *The Markov decision process*

Reinforcement learning (RL) is a machine learning algorithm that sorts out the Markov decision process (MDP) in the formula of a tuple:{*S, A, T, R*}, where *S* is a set of environment states, *A* is a set of actions, *T* is a transition function, *R* is a set of reward or losses. An agent in an MDP environment would learn how to take action from *A* by observing the environment with states from *S*, according to corresponding transition probability *T* and reward *R* achieved from the interaction. The Markov property indicates that the state transitions are only dependent on the current state and current action is taken, but independent to all prior states and actions[52]. As known that a textile manufacturing process has a number of parameter variables as *P* {$v_1, v_2... v_n$}, if the probable value of $v_j$ is $p(v_j)$, the parameter of the process defining the environment space $\varphi$ from $\prod_{j=1}^{n} p(v_j), v_j \in V_j$ impacting the performance of textile process with regards to criteria {$c_1, c_2... c_m$}. These parameter variables are independent to each other and obey a Markov process that models the stochastic transitions from a state $S_t$ at time step *t* to next state $S_{t+1}$, where the environment state at time step *t* is:

$$S_t = [s_t^{v_1}, s_t^{v_2} ... s_t^{v_n}] \in \varphi \tag{10}$$

RL trains an agent to act optimally in a given environment based on the observation of states and the feedback from their interaction, acquiring rewards and maximizing the accumulative future rewards over time from the interaction[52]. Here, the agent learns in the interaction with the environment by taking actions that can be conducted on the parameter variables $\in P$ {$v_1, v_2... v_n$} at time step *t*. More specifically, in a time step *t*, the action of each single variable $v_j$ could be kept (0) or changed up (+) / down (-) in the given range with specific unit $u_j$. So there are $3^n$ actions totally in the action space and, for simplicity, the action vector $A_t$ at time step *t* could be:

$$A_t = [a_t^{v_1}, a_t^{v_2} ... a_t^{v_n}], \quad where \; a_t^{v_j} \in \{-u_j, 0, +u_j\}, v_j \in V_j. \tag{11}$$

The state transition probabilities, as mentions that, are only dependent on the current state $S_t$ and action $A_t$. It specifies how the reinforcement agent takes action $A_t$ at time step *t* to transit from $S_t$ to next state $S_{t+1}$ in terms of *T* ($S_{t+1} | S_t, A_t$). For all $a_t^{v_j} \in \{-u_j, 0, +u_j\}, v_j \in V_j$, $T(S_{t+1} | S_t, A_t) > 0$ and $\sum_{S_{t+1} \in \varphi} T(S_{t+1} | S_t, A_t) = 1$. The reward achieved by an agent in an environment is specifically related to its transition between states, which evaluates how good the transition agent conducts and facilitates the agent to converging faster to an optimal solution.

### 3.5. *Deep-Q-network algorithm*

The RL performs a vital function in the MDP problem. However, the basic RL algorithms in most of the studies, such as the Q-learning and the SARSA (0/λ), are based on a memory-intensive tabular representation (i.e. Q-table) of the value, or instant reward, of taking an action *a* in a specific state *s* (the Q value of state-action pair, a.k.a Q(s, a)). The tabular algorithms would restrict the application of the RL in realistic large-scale cases when the amounts of states or actions are tremendous. Because in these situations, not only the tables come short of recording all of the Q(s,a), but presenting computational power would be overwhelming as well.

The deep neural networks (DNNs) is another widely applied machine learning technique that is quite good at coping with the large-scale issues and has recently been combined with the RL to evolve toward deep reinforcement learning (DRL). Deep-Q-network is a DRL developed by Mnih et al[53] in 2015 as the first artificial agent that is capable of learning policies directly from high-dimensional sensory inputs and agent-environment interactions. It is an RL algorithm proposed based on Q-learning which is one of the most widely used model-free off-policy and value-based RL algorithms.

### 3.5.1. Q-learning

Q-learning learns through estimating the sum of rewards $r$ for each state $S_t$ when a particular policy $\pi$ is being performed. It uses a tabular representation of the $Q^\pi(S_t, A_t)$ value to assign the discounted future reward $r$ of state-action pair at time step $t$ in Q-table. The target of the agent is to maximize accumulated future rewards to reinforce good behavior and optimize the results. In Q-learning algorithm, the maximum achievable $Q^\pi(S_t, A_t)$ obeys Bellman equation on the basis of an intuition: if the optimal value $Q^\pi(S_{t+1}, A_{t+1})$ of all feasible actions $A_{t+1}$ on state $S_{t+1}$ at the next time step is known, then the optimal strategy is to select the action $A_{t+1}$ maximizing the expected value of $r + \gamma \cdot max_{A_{t+1}} Q^\pi(S_{t+1}, A_{t+1})$.

$$Q^\pi(S_t, A_t) = r + \gamma \cdot max_{A_{t+1}} Q^\pi(S_{t+1}, A_{t+1}) \qquad (12)$$

According to the Bellman equation, the Q-value of the corresponding cell in Q-table is updated iteratively by:

$$Q^\pi(S_t, A_t) \leftarrow Q^\pi(S_t, A_t) + \alpha [r + \gamma \cdot max_{A_{t+1}} Q^\pi(S_{t+1}, A_{t+1}) - Q^\pi(S_t, A_t)] \qquad (13)$$

where $S_t$ and $A_t$ are the current state and action respectively, while $S_{t+1}$ is the state achieved when executing $A_{t+1}$ in the set of $S$ and $A$ in any given MDP tuples of $\{S, A, T, R\}$. $\alpha \in [0, 1]$ is the learning rate, which indicates how much the agent learned from new decision-making experience ($Q^\pi(S_{t+1}, A_{t+1})$) would override the old memory ($Q^\pi(S_t, A_t)$). $r$ is the immediate reward, $\gamma \in [0, 1]$ is the discount factor determining the agent's horizon.

The agent takes action on a state in the environment and the environment interactively transmits the agent to a new state with a reward signal feedback. The basic principle of Q-learning RL essentially relies on a trial and error process, but different from humans and other animals who tackle the real-world complexity with a harmonious combination of RF and hierarchical sensory processing systems, the tabular representation of Q-learning is not efficient at presenting an environment from high-dimensional inputs to generalize past experience to new situations[53].

### 3.5.2. DQN: innovative combination of deep neural networks and Q-learning

Q-table saves the Q value of every state coupled with all its feasible actions in an environment, while the growing complexity in the problem nowadays indicates that the states and actions in an RL environment could be innumerable (such as Go game). In this regard, DQN applies DNNs instead of Q-table to approximate the optimal action-value function. The DNNs feed by the state for approximating the Q-value vector of all potential actions, for example, are trained and updated by the difference between Q-value derived from previous experience and the discounted reward obtained from the current state. While more importantly, to deal with the instability of RL representing the Q value using nonlinear function approximator[54], DQN innovatively proposed two ideas termed experience replay[55] and fixed Q-target. As known that Q-learning is an off-policy RL, it can learn from the current as well as prior states. Experience replay of DQN is a biologically inspired mechanism that learns from randomly taken historical data for updating in each time step, which therefore would remove correlation in the observation sequence and smooth over changes in the data distribution. Fixed Q-target performs a similar function, but differently, it reduces the correlations between the Q-value and the target by using an iterative update that adjusts the Q-value towards target values periodically.

Specifically, the DNNs approximate Q-value function in terms of $Q\text{-}(s, a; \theta_i)$ with parameters $\theta_i$ which denotes weights of Q-networks at iteration $i$. The implementation of experience replay is to store the agent's experiences $e_t = (S_t, A_t, r_t, S_{t+1},)$ at each time step $t$ in a dataset $D_t = \{e_1, \ldots e_t,\}$. Q-learning updates were used during learning to samples of experience, (S, A, r, S') ~ U(D), drawn uniformly at random from the pool of stored samples. The loss function of Q-networks update at iteration *i* is:

$$L_i(\theta_i) = \mathbb{E}_{(S,A,r,S') \sim U(D)}\left[\left((r + \gamma \cdot \max_{A'} Q(S', A'; \theta_i^-) - Q(S, A; \theta_i)\right)^2\right] \quad (14)$$

where $\theta_i^-$ are the networks weights from some previous iteration. The targets here are dependent on the network weights; they are fixed before leaning begins. More precisely, the parameters $\theta_i^-$ from the previous iteration are fixed as optimizing the $i_{th}$ loss function $L_i(\theta_i)$ at each stage and are only updated with $\theta_i$ every *R* steps. To implement this mechanism, DQN uses two structurally identical but parametrically differential networks, one of it predicts $Q(S, A; \theta_i)$ using the new parameters $\theta_i$, the rest one predicts $r + \gamma \cdot \max_{A'} Q(S', A'; \theta_i^-)$ using previous parameters $\theta_i^-$. Every *R* steps, the *Q* network would be cloned to obtain a target network $\hat{Q}$, and then $\hat{Q}$ would be used to generate Q-learning target $r + \gamma \cdot \max_{A'} Q(S', A'; \theta_i^-)$ for the following *R* updates to network *Q*.

The DQN is a typical and classic DRL algorithm that played a key role in the applications of production scheduling, playing video games, and the Computer Go [31], [53], [56], etc. Given the advantages that the DQN can offer when confronted with decision making in textile process optimization, this algorithm would be adopted in the construction of the proposed decision support system.

## 4. System framework

Figure 1 illustrates the textile manufacturing process optimization problem in the paradigm of RL, where the decision-maker plays the role of agent to traverse and explore the state space, the environment includes all the targeted process parameter variables, the adjustment of parameter variables indicates the action, the solution combined of all the parameter variables represents the state, and the objective function denotes the reward. The objective of the developed decision support system is to optimize the textile process with regards to its parameter variables on the basis of the multi-criteria objective function which has fundamentally been formulated in Equation(3). Therefore, here the feedback from the environment depending on a reward function is in accordance with the objective function.

Machine learning library of Scikit-learn is employed to develop the RF models [57], where the sub-sample size of it is always the same as the original input sample size but the samples are drawn with a replacement if *bootstrap* is used (or else the whole dataset would be used to build each tree). RF algorithm starts from randomly drawing a number of samples from the original data and grows corresponding regression trees (*n_estimators*) in accordance with the drawn samples. In each leaf node of regression tree, respectively given the minimum number of samples required to be at a leaf node (*min_samples_leaf*) and to split an internal node (*min_samples_split*) as well as the maximum depth of the tree (*max_depth*), a number of the best splits (*max_features*) randomly selected from all variables are considered for binary partitioning. The selection of the feature for node splitting from a random set of features decreases the correlation between different trees and thus the average prediction of multiple regression trees is expected to have lower variance than individual regression trees[47]. A hyperparameter tuning process using a grid search with 3-fold cross-validation is conducted in this study for optimizing the RF models, and the applied parameter grid with 3960 combinations is demonstrated in Table 1.

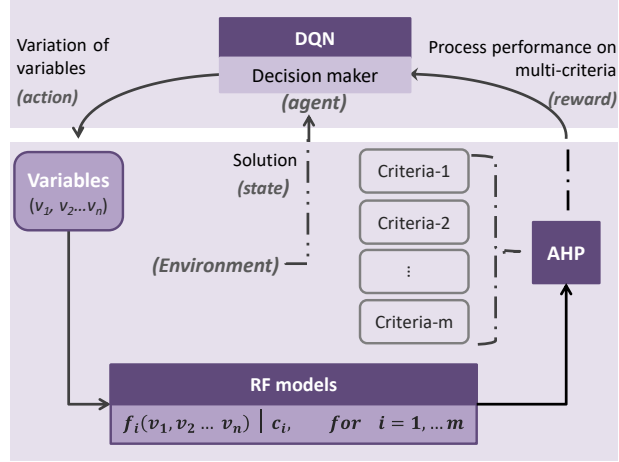

Figure 1. The MDP structure of textile manufacturing process multi-criteria optimization in the proposed framework

TABLE I. PARAMETER GRID USED IN HYPERPARAMETER TUNING PROCESS

| Parameter of RF | Options and implication |
|---|---|
| bootstrap | True or False : sampling data points with or without replacement |
| n_estimators | 200, 400, 600, 800, 1000, 1200, 1400, 1600, 1800, 2000 |
| min_samples_leaf | 1, 2, 4 |
| min_samples_split | 2, 5, 10 |
| max_depth | 10, 20, 30, 40, 50, 60, 70, 80, 90, 100, None |
| max_features | 'auto': $max\_features = x$ (the number of observed input vector) <br> 'sqrt': $max\_features = \sqrt{x}$ |

---

**Algorithm 1: Multi-criteria decision support system main body:**

**Initialize:** $D_e$ (process experience data), $C_m$ (comparison matrix of considered criteria),
  $P$ ($p_1, p_2 \dots p_m$, expected performance of process), $E$ (number of episodes), $N$ (number of time steps),
  $\alpha$ (learning rate), $\gamma$ (discount factor), $R$ (the step updating DQN), $D$ (replay memory size);

**Phase 1: RF model construction**
Split input and output of $D_e$ to train and test RF models respectively;
**For** each output (process performance in regard to criterion $c_i$) **do**
  RF model $f_i(v_1, v_2 \dots v_n) \mid c_i \leftarrow parameter\ tuning\ result\ with\ min\ error\ (see\ TABLE\ 1)$
**End for**

**Phase 2: Multi-criteria summary**

Initialize and check multi-criteria pairwise comparison matrix $C_m = \begin{bmatrix} 1 & \cdots & a_{1m} \\ \vdots & \ddots & \vdots \\ a_{m1} & \cdots & 1 \end{bmatrix}$;

Geometric mean $GM_i = \{\prod_{j=1}^{m} a_{ij}\}^{\frac{1}{m}}$ and relative weight of criterion $w_i = \frac{GM_i}{\sum_{i=1}^{m} GM_i}$;

Transformation: $argmax_{v_j \in V_j} [f(v_1, v_2 \dots v_n) \mid C] = argmax_{v_j \in V_j} \sum_{i=1}^{m} w_i f_i(v_1, v_2 \dots v_n)$;

**Phase 3: Optimization using DQN**
Initial function $Q$ with random weights $\theta$:
Initial function $\hat{Q}$ with weights $\theta^- = \theta$;
Initialize state $s_0 = (v_1, v_2 \dots v_n)$
**For** episode =1, $E$ **do**
  **For** time step=1, $N$ **do**

Choose an action $a_t$ using ε-greedy policy
   Execute action $a_t$, observe next state $s_{t+1}$
   Estimate $f(s_t)$ and $f(s_{t+1})$ to observe $r_t$ ($r_t = \sum_{i=1}^{m} \sqrt{w_i^2(f_i(s_t) - p_i)^2} - \sum_{i=1}^{m} \sqrt{w_i^2(f_i(s_{t+1}) - p_i)^2}$)
   Store transition ($s_t$, $a_t$, $r_t$, $s_{t+1}$) in $D$
   Sample random minibatch of transitions ($s_t$, $a_t$, $r_t$, $s_{t+1}$) from $D$
   Set $y_i = \begin{cases} r_j & \text{if terminates at step } j+1 \\ r_j + \gamma \max_{a'} \hat{Q}(s_{j+1}, a'; \theta^-) & \text{otherwise} \end{cases}$
   Perform a gradient descent step on $\left(y_i - Q(s_j, a_j; \theta)\right)^2$ with regard to $\theta$
   Every $R$ steps reset $\hat{Q} = Q$
   $s_t \leftarrow s_{t+1}$
   **End For**
**End For**

The pseudo-code of the proposed multi-criteria decision support system based on DQN reinforcement learning, RF models, and AHP are illustrated in Algorithm 1, and an episodic running within the algorithm is graphically displayed in Figure 2. Apart from the aforementioned parameters, it is also necessary to provide an experience dataset ($D_e$) regarding the textile process modeling and the expected process performance or optimization targets ($P$) to the system construction. In order to balance the exploration and exploitation of states at the learning period and optimizing period respectively, we initialize the first state of every episode randomly from each sub-state $s_t^{v_i}$ where parameter variables $v_j \in V_j$, and more importantly, apply increasing ε-greedy policy at the meantime.

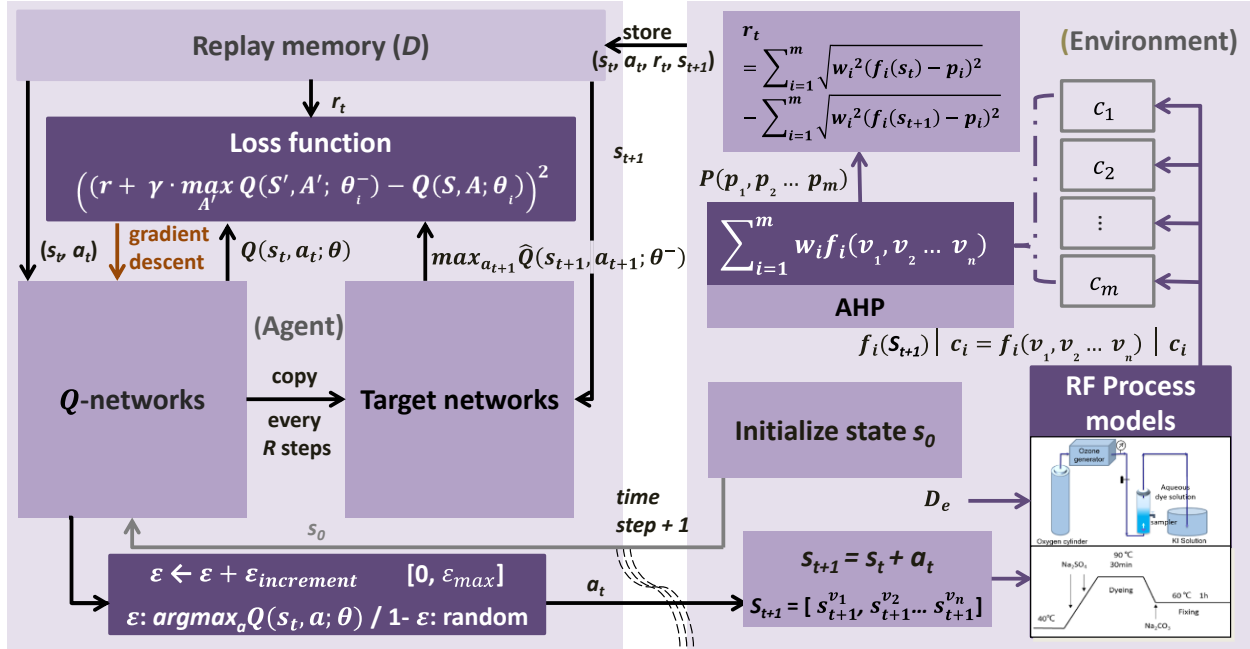

Figure 2. Flowchart of the algorithm implementing the proposed deep reinforcement learning based multi-criteria decision support system for textile manufacturing process optimization

**Algorithm 2:** Choose an action using ε-greedy policy

**Input:** $\varepsilon_{increment}, \varepsilon_{max}$

$\varepsilon \leftarrow \varepsilon + \varepsilon_{increment}$ $(0 \leq \varepsilon \leq \varepsilon_{max})$;

**If** random(0,1) > $\varepsilon$

   Randomly choose action $a_t$ from action space

**Else**

   Select $a_t = argmax_a Q(s_t, a; \theta)$

**End if**

In fact, the algorithm given above can work without episodes, as the target of an RL trained agent is to find the optimized solution, in terms of state in the environment with minimum error tested by RF models and AHP evaluations, however, the lack of exploration of the agent in an environment may cause local optimum in a single running. So we initialize the first state randomly and introduce an episodic learning process to the agent for enlarging the exploration and preventing local optimum. On the other hand, we apply the increasing $\varepsilon$-greedy policy as well. The $\varepsilon$-greedy policy helps the agent find the best action (maximum Q value) in the present state to go to the next state with a possibility of $\varepsilon$ that may also randomly choose an action with a possibility of $1-\varepsilon$ to get a random next state. While, as illustrated in Algorithm 2, increasing $\varepsilon$-greedy is employed with an increment given in each time step from 0 until it equals to $\varepsilon_{max}$. This benefits the agent to explore the unexplored states without staying in the exploitation of already experienced states of Q-networks, and plentifully exploit them when the states are traversed enough.

## 5. Case study

The application of the established decision support system to a textile color fading ozonation process was performed to evaluate the performance of the proposed framework. Color fading is an essential textile finishing process for specific textile products such as denim to obtain a worn fashion style[17], but it conventionally was achieved by chemical methods that have a high cost and water consumption, as well as a heavy negative impact on the environment. Instead, ozone treatment is an advanced finishing process employing ozone gas to achieve color faded denim without a water bath, which consequently causes less environmental issues. The complexities of this process have been investigated in our previous works[58]–[61], and according to the experience data with 129 samples we collected from that application, the setup of the proposed framework would be attempted to solve a 4-targets optimization problem of color fading ozonation process.

In terms of the experience dataset used to train and test RF models, it includes 4 process parameters (water-content, temperature, *pH* and time) of the process and 4 process performance index known as *k/s*, $L^*$, $a^*$, and $b^*$ of the treated fabrics. Where the *k/s* value indicates color depth, while $L^*$, $a^*$, and $b^*$ illustrate the color variation in three dimensions (lightness, chromatic component from green to red and from blue to yellow respectively). Normally, the color of the final textile product in line with specific *k/s*, $L^*$, $a^*$, and $b^*$ is within the acceptable tolerance of the consumer.

### 5.1. *Modeling color fading ozonation process using the random forest*

In terms of the RF model construction, 75% of the data was divided into the training group and the rest 25% was used to test models. In order to decrease the bias and promote the generalization of applied RF models in the system, we have trained 4 separate models for predicting 4 outputs (*k/s*, $L^*$, $a^*$, and $b^*$) respectively. The results of hyperparameter tuning in regards with the context given in Table 1 are displayed in Table 2, and the final optimized models are tested (25% with unseen data) that can well predict the process performance with accuracies (R-square) of 0.996, 0.954, 0.937 and 0.965 respectively.

TABLE II. THE RF MODELS HYPERPARAMETER TUNING RESULT AND THE TESTING RESULTS OF OPTIMIZED MODEL

| *Bootstrap* | *n_* | *Min_* | *Min_* | *Max_* | *Max_* | $R^2$ | *MAE* | *MAPE* |
|---|---|---|---|---|---|---|---|---|

|       | estimators | samples_leaf | samples_split | depth | eatures |       |      |       |
|-------|------------|--------------|---------------|-------|---------|-------|------|-------|
| k/s   | True       | 2000         | 1             | 2     | 30      | 'auto' | 0.996 | 0.28 | 12.36 |
| $L^*$ | False      | 2000         | 1             | 2     | None    | 'sqrt' | 0.954 | 0.77 | 8.99  |
| $a^*$ | False      | 2000         | 2             | 2     | None    | 'auto' | 0.937 | 2.29 | 15.43 |
| $b^*$ | True       | 2000         | 1             | 5     | 100     | 'auto' | 0.965 | 2.87 | 11.62 |

TABLE III. PAIRWISE COMPARISON MATRIX OF K/S, L*, A* AND B* WITH RESPECT TO THE OVERALL COLOR PERFORMANCE

|       | k/s  | $L^*$ | $a^*$ | $b^*$ | GM     | w     |
|-------|------|-------|-------|-------|--------|-------|
| k/s   | 1    | 3     | 5     | 5     | 2.9428 | 0.556 |
| $L^*$ | 1/3  | 1     | 3     | 3     | 1.3161 | 0.249 |
| $a^*$ | 1/5  | 1/3   | 1     | 2     | 0.6043 | 0.114 |
| $b^*$ | 1/5  | 1/3   | 1/2   | 1     | 0.4273 | 0.081 |

### 5.2. *Determining the criteria weights using the analytic hierarchy process*

By means of combining experts' judgment with our experience, a pairwise comparison matrix of the 4 decision criteria with respect to the overall color performance of ozonation process treated textile product is provided in Table 3. $\lambda_{max}$ of this comparison matrix is 4.1042 and known that the *RCI* for 4 criteria problem is 0.90, as a result, the *CR* calculated is 0.0386≤0.08 which implies that the evaluation within the matrix is acceptable.

### 5.3. *Deep Q-Networks for optimal decision-making*

We optimize the color performance in terms of *k/s*, $L^*$, $a^*$, and $b^*$ of the textile in ozonation process by finding a solution including proper parameter variables of water-content, temperature, *pH* and treating time that minimizes the difference between such specific process treated textile product and the targeted sample. Therefore, the state space $\varphi$ in this case is composed by the solutions with four parameters (water-content, temperature, *pH* and treating time) in terms of $S_t = [s_t^{v_1}, s_t^{v_2}, s_t^{v_3}, s_t^{v_4}]$. In a time step *t*, the adjustable units of these parameter variables are 50, 10, 1 and 1 respectively in the range of [0, 150], [0,100], [1, 14] and [1, 60] respectively. As the action of a single variable $v_j$ could be kept (0) or changed up (+) / down (-) in the given range with specific unit *u*, so there are $3^4 = 81$ actions totally in the action space and the action vector at time step *t* is $A_t = [a_t^{v_1}, a_t^{v_2}, a_t^{v_3}, a_t^{v_4}]$, where $a_t^{v_1} \in \{-50, 0, +50\}, v_1 \in [0, 150]$ ; $a_t^{v_2} \in \{-10, 0, +10\}, v_2 \in [0, 100]$ ; $a_t^{v_3} \in \{-1, 0, +1\}, v_3 \in [1, 14]$ ; $a_t^{v_4} \in \{-1, 0, +1\}, v_4 \in [1, 60]$.

The transition probability is 1for the states in the given range of state space above, but 0 for the states out of it. The reward *r* at time step *t* is expected to be in line with how close the agent gets to our target, and as the relative importance of these four performance criteria (0.556, 0.249, 0.114 and 0.081 respectively) is analyzed in AHP, we could set up the reward function as illustrated below to induce the agent to approach our optimization results:

$$r_t = \sum_{i=1}^m \sqrt{w_i^2(f_i(s_t) - p_i)^2} - \sum_{i=1}^m \sqrt{w_i^2(f_i(s_{t+1}) - p_i)^2} \tag{15}$$

As shown the pseudo-code of DQN main body in Algorithm 1, optimization targets of textile ozonation process are needed to function the system ($p_1$, $p_2$, $p_3$, $p_4$, the color performance of the ozonation process in terms of *k/s*, $L^*$, $a^*$, and $b^*$), these targets in the present case study would be sampled by experts. In addition to the targets, the parameters of DQN such as step *R* for updating Q-networks and replay memory size *D*, as well as the learning rate $\alpha$ and the discount rate $\gamma$ for updating loss function, etc., are listed in Table 4. In particular, the *R* step for updating DQN here denotes that after 100 steps, the Q-networks would be updated at every 5 steps.

TABLE IV. DQN ALGORITHM SETTING IN TEXTILE OZONATION PROCESS CASE STUDY

| R       | D    | $\alpha$ | $\gamma$ | $\varepsilon_{increment}$ | $\varepsilon_{max}$ | E | N    |
|---------|------|----------|----------|---------------------------|---------------------|---|------|
| 5(>100) | 2000 | 0.01     | 0.9      | 0.001                     | 0.9                 | 5 | 5000 |

## 5.4. *Results and discussion*

As mentioned that there are four RF models trained for predicting $p_1$, $p_2$, $p_3$, $p_4$ of the color performance of the ozonation process in terms of *k/s*, $L^*$, $a^*$ and $b^*$, respectively. The predictive performance of these models displayed in Figure 3 clearly indicates that the models work steadily in the algorithm of present case study as it is found that the models predicted values are generally in accordance with the actual measured values through the predictive errors of different models varied slightly at different levels. This finding furthermore reflects that the RF approach is capable of modeling the textile manufacturing process and plays a significant role in our proposed decision support system. It is worth remarking that the average error of MAPE is higher than 10%. This is due to the fact that the targets in these models, i.e. *k/s*, $L^*$, $a^*$, and $b^*$, are the colorimetric values ranging from -120 to 120, while the sample data applied are close to 0 and mostly are negative, which exaggerated the evaluation of MAPE in this issue.

In particular, the neural networks implemented by TensorFlow[62] are used in our case study to realize Q-networks which have been described in detail in the developed framework of Algorithm 1. The networks consist of two layers with 50 and $3^4$ hidden nodes respectively, where the last layer corresponds to the actions. As demonstrated in Table 5, there are 5 experimental targets sampled by experts that were used in the present case study.

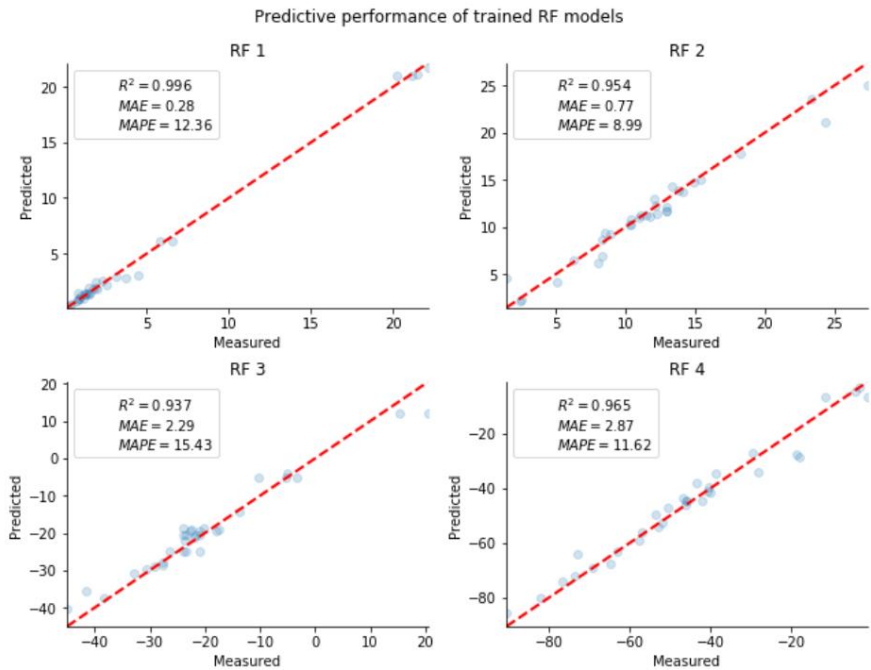

Figure 3. Predictive performance of the RF models trained in the case study for supporting decision making in textile color ozonation process

TABLE V. THE EXPERIMENTAL TARGETS SAMPLED BY EXPERTS WE USED IN THE CASE STUDY APPLICATION OF PROPOSED DECISION SUPPORT SYSTEM

|       | 1      | 2      | 3      | 4      | 5      |
|-------|--------|--------|--------|--------|--------|
| *k/s* | 0.81   | 1.00   | 2.45   | 1.84   | 0.41   |
| $L^*$ | 15.76  | 11.63  | 8.2    | 9.72   | 21.6   |
| $a^*$ | -20.84 | -24.08 | -18.73 | -21.09 | -36.48 |
| $b^*$ | -70.79 | -54.1  | -38.17 | -42.78 | -59.95 |

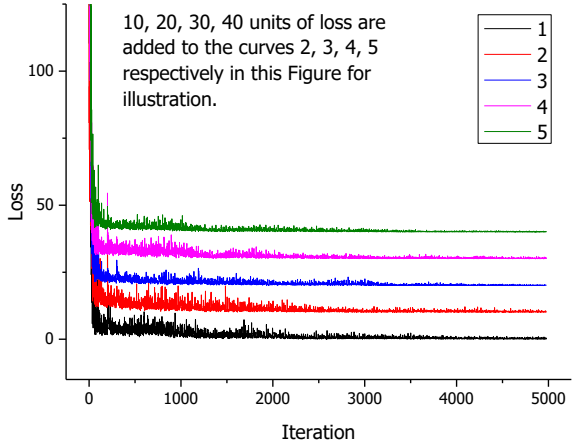
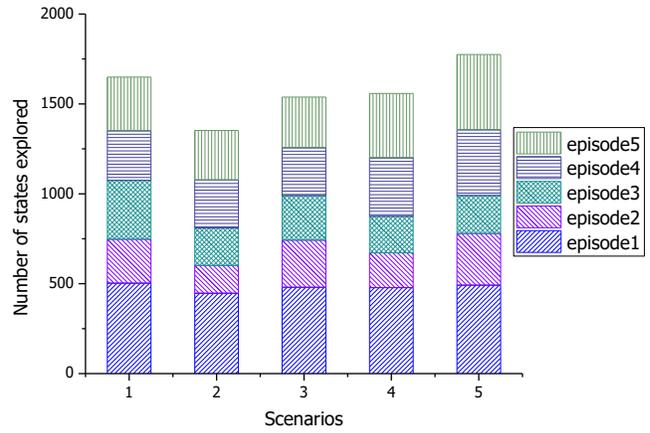

Figure 4. The loss function of target networks for each scenario with different targets

Figure 5. The number of states explored by the DQN agent in each episode for different scenarios

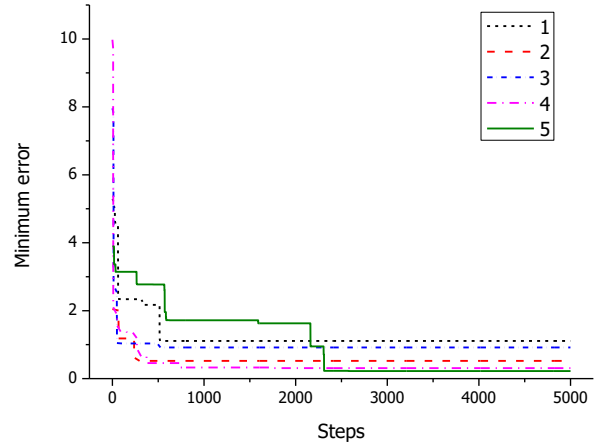

Figure 6. The minimum error of solutions that DQN agent has achieved along with the time steps

Figure 4 demonstrates the loss function of target Q-networks for each scenario. It is found that converged quickly to be steady after training by the action values feedback obtained from the environment in early time steps, which denotes that the representation of Q-value in this Q networks is stable and accurate. Relating to the dramatic falls at the beginning steps in each scenario, it is owing to the increasing $\varepsilon$-greedy policy employed in the algorithm which leads the agent to choose action randomly with a high probability in a range of beginning time steps, but increasingly choose the action with high value after that. On the other hand, in terms of the 5 episodes we employed in each scenario, it is found the maximum unrepeated states that a DQN agent has been explored that are all occurred in the first episode in all the scenarios (Figure 5). This also reflects that the increasing $\varepsilon$ has balanced the process of exploration and exploitation of states in the environment, and the rest episodes would benefit from the experience achieved before.

This case study implemented 5 episodic trails for each scenario with different targets in the proposed framework. We collect the minimum error of solutions (states) evaluated by RF-AHP during the DQN agent interacted with the environment time steps. While as the initial state is randomly given in our proposed system for avoiding local optimum, only the ones with the best results are specifically illustrated in Figure 6. regarding the time steps. It is found that the reward function can effectively guide the agent to find the optimum in the environment, and the times steps taken in 3000 seem enough for the optimization. However, it is worth noting that the efficiency of the reward function in our proposed decision support system is still not fully illustrated by this case study. One main reason for this comes to the limited data of textile manufacturing processes and the costly computational power which hopefully would be solved in the Industry 4.0 era.

TABLE VI. SIMULATED RESULTS OF SOLUTIONS WITH MINIMUM ERRORS OBTAINED FROM DQN BASED AND Q-LEANRING BASED FRAMEWORK RESPECTIVELY

|  | 1 | 2 | 3 | 4 | 5 |
|---|---|---|---|---|---|
| DQN | 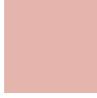 | 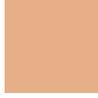 | 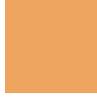 | 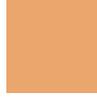 | 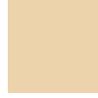 |
| Targets | 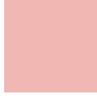 | 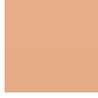 | 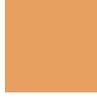 | 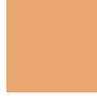 | 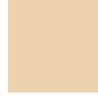 |
| Q-learning | 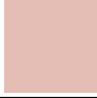 | 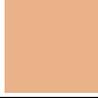 | 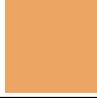 | 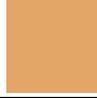 | 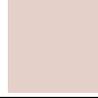 |

In order to show the advantage and effectiveness of DQN in our proposed decision support system, a comparison with Q-learning based on the same developed framework is conducted, and the simulated color performance of the results in terms of the solutions with minimum error obtained from two methods are comparatively demonstrated in Table 6 with the targets. Here the error is calculated by Equation (16), which are 1.06, 0.50, 0.88, 0.29, 0.22 and 1.13, 0.54, 0.91, 1.28, 1.76 in the DQN and Q-learning based decision support system for scenarios from 1 to 5 respectively.

$$error = \sqrt{0.556^2(k/s_s - k/s_t)^2 + 0.249^2(L_s^* - L_t^*)^2 + 0.114^2(a_s^* - a_t^*)^2 + 0.081^2(b_s^* - b_t^*)^2} \quad (16)$$

where $k/s_s$, $L_s^*$, $a_s^*$, $b_s^*$ are the properties of simulated color performance of solution obtained from the decision support system, and $k/s_t$, $L_t^*$, $a_t^*$, $b_t^*$ are the targeted color performance.

## 6. Limitation and future prospective

The proposed decision support system is clearly applicable in the textile manufacturing sector, which has been tested in the implemented case study. It is worth mentioning that this system can also be applied in practice with different objective functions such as energy optimization, material optimization, etc. However, it is worth remarking that certain features of this framework may hinder the massive promotion and application of it. The AHP performs well in the MCDM problems, while it relies heavily on experts' estimation, which may limit its application in certain areas[63], [64]. Meanwhile, it is well known that the practice and effectiveness of RF and DQN rely strongly on big data and computation power, which is quite limited in the application of the textile industry nowadays. But the application of artificial intelligent techniques is growing in the textile manufacturing industry, such concerns could be properly addressed in the industry 4.0 era when it is able to take full advantage of the Internet of Things (IoT) environment. Future development of the proposed decision support system should be able to learn from the interaction with the complexity-growing environment online by constantly feeding new data and scenarios to follow the development of the textile manufacturing process. Future research should devote more effort to test the proposed framework and broaden the application of it in more textile manufacturing processes by collecting real empirical data and construct the corresponding MDP paradigm.

## 7. Conclusions

Textile manufacturing is a traditional industry involving high complexities in interconnected processes with limited capacity on the application of modern technologies. Decision-making in this domain generally takes multiple criteria into consideration, which usually arouses more complexity. Traditional classical approaches are no longer efficient owing to the growing complexity with large-scale data and high dimensional decision space in some scenarios. In this paper, a decision support system combining the random forest model, analytical hierarchy process and deep Q-Networks is proposed for optimizing the textile manufacturing process. This developed system tackles large scale optimization problems in high dimensional decision space with multi-criteria in the textile manufacturing process. Empirical data and human knowledge of the textile process are needed to build random forest models and evaluate criteria respectively. The dependence of the operations on data and knowledge of this system are in accordance with the characteristics of the complicated textile manufacturing process with respect to both objective and subjective factors in decision making of an application. This system describes decision making for optimizing the textile manufacturing process in the Markov decision process paradigm of {*S, A, T, R*} in the proposed algorithm, and takes advantage of the deep reinforcement learning by means of a deep Q-Networks algorithm in this proposed framework to exploit the data on the basis of random forest models and AHP multi-criteria structure to find the optimal textile manufacturing. The application in optimizing a textile ozonation process was released. The results showed that the developed system is capable of learning to master the challenging decision-making tasks and performed better than traditional methods.

**Acknowledgments**

This research was supported by the funds from National Key R&D Program of China (Project NO: 2019YFB1706300), and Scientific Research Project of Hubei Provincial Department of Education, China (Project NO: Q20191707).

The first author would like to express his gratitude to China Scholarship Council for supporting this study (CSC, Project NO. 201708420166).